\pdfoutput=1

\documentclass[11pt]{article}

\usepackage[final]{ACL2023}

\usepackage{times}
\usepackage{latexsym}
\usepackage{multirow}
\usepackage{todonotes}
\usepackage{amsmath}

\newtheorem{problem}{Problem}

\usepackage{enumitem}

\usepackage[T1]{fontenc}

\usepackage[utf8]{inputenc}
\usepackage{graphicx} 
\usepackage{xcolor}
\usepackage{courier}  
\usepackage{booktabs}
\usepackage{hyperref}
\usepackage{amssymb}
\usepackage{amsmath}
\usepackage{latexsym}
\usepackage{graphicx}
\usepackage{xcolor}
\usepackage{soul}
\sethlcolor{lightgray}
\usepackage{booktabs}
\DeclareMathOperator*{\argmax}{arg\,max}

\usepackage{microtype}
\usepackage{pifont}
\newcommand{\cmark}{\ding{51}}%
\newcommand{\xmark}{\ding{55}}%

\author{Nancy Tyagi \quad Aidin Shiri \quad Surjodeep Sarkar \quad
  \textbf{Abhishek Kumar Umrawal} \quad \textbf{Manas Gaur} \\
  University of Maryland, Baltimore County \\
  \texttt{\{nancyt1, aidins1, ssarkar1, aumrawal, manas\}@umbc.edu}}

\title{\emph{Simple is Better and Large is Not Enough}:\\ Towards Ensembling of Foundational Language Models}

\date{}

\begin{document}
\maketitle
\begin{abstract}

Foundational Language Models (FLMs) have advanced natural language processing (NLP) research. Current researchers are developing larger FLMs (e.g., XLNet, T5) to enable contextualized language representation, classification, and generation. While developing larger FLMs has been of significant advantage, it is also a liability concerning hallucination and predictive uncertainty. Fundamentally, larger FLMs are built on the same foundations as smaller FLMs (e.g., BERT); hence, one must recognize the potential of smaller FLMs which can be realized through an ensemble. In the current research, we perform a reality check on FLMs and their ensemble on benchmark and real-world datasets. We hypothesize that the ensembling of FLMs can influence the individualistic attention of FLMs and unravel the strength of coordination and cooperation of different FLMs. We utilize BERT and define three other ensemble techniques: \{Shallow, Semi, and Deep\}, wherein the Deep-Ensemble introduces a knowledge-guided reinforcement learning approach. We discovered that the suggested Deep-Ensemble BERT outperforms its large variation i.e. $BERT_{large}$, by a factor of many times using datasets that show the usefulness of NLP in sensitive fields, such as mental health. 


\end{abstract}

\section{Introduction} \label{sec:introduction}
Foundational language models (FLMs) have raised unreasonable expectations in achieving human-level performance on various classification, generation, and representation learning tasks \cite{tamkin2021understanding}. However, the methodology of the behavioral testing of NLP models by \cite{ribeiro2020beyond} and the recent HELM\footnote{Holistic Evaluation of Language Models} did motivate the community to rethink the purpose and effectiveness of FLMs \cite{liang2022holistic}. For instance, it is easy for an FLM to tell whether \textbf{Sentence 1}: \textit{Patient also required multiple pressors to maintain her blood pressure.} entails/contradicts \textbf{Sentence 2}: \textit{the patient is hypertensive} or possesses a neutral relationship; however, if we shift the domain and ask the same FLM to provide a Yes/No label to the following. The question ``Does the person have Depression?'' given context: ``What I have is depression even though manic episodes aren't characteristic of depression.'' the model fails by giving ``Yes'' as an outcome. 

The critical insight from \cite{devlin2018bert} was to train deep learning models to focus on essential features and adjust attention weights accordingly. This insight enabled practitioners to prepare FLMs on massive training data to achieve generalization across multiple tasks. \cite{bommasani2021opportunities} describes the emergent capabilities of large FLMs but also mentions that the quality and consistency in the model's performance do matter. Several approaches, such as probing with auxiliary tasks \cite{chen2022probing}, knowledge distillation \cite{jiao2019tinybert}, and model cascades \cite{wang2020wisdom}, have been proposed to allow small-sized FLMs to take control where larger FLMs fall short of providing acceptable outcomes. 

\begin{table*}[!ht]
\scriptsize
\centering
\begin{tabular}{lccccc}

\toprule
\multicolumn{1}{c}{Text Sentence}                                                                                                                 & Expected      & $BERT_{tiny}$         & $BERT_{Mini}$        & $BERT_{Base}$         & $BERT_{Large}$        \\ \toprule
Give a listen to Gramatik, he is \hl{amazing}.                                                                                                         & Admiration    & \multirow{2}{*}{\xmark } & \multirow{2}{*}{\xmark } & \multirow{2}{*}{\xmark } & \multirow{2}{*}{\xmark } \\   \\ \hline
\begin{tabular}[c]{@{}l@{}}Wish more people shared your thoughts,\\ it will make the world a \hl{better} place.\end{tabular}                           & Caring        & \multirow{2}{*}{\cmark } & \multirow{2}{*}{\xmark } & \multirow{2}{*}{\xmark } & \multirow{2}{*}{\xmark } \\ 
  \\ \hline
I'm \hl{disappointed} in you too.                                                                                                                      & Sadness       & \multirow{2}{*}{\xmark} & \multirow{2}{*}{\cmark} & \multirow{2}{*}{\xmark} & \multirow{2}{*}{\xmark} \\
 \\ \hline
\begin{tabular}[c]{@{}l@{}}FYI: Time in Europe works the same as the State. \\ We do not have '\hl{weird} time'.\end{tabular}                           & Annoyance     & \multirow{2}{*}{\xmark} & \multirow{2}{*}{\xmark} & \multirow{2}{*}{\cmark} & \multirow{2}{*}{\xmark} \\
            \\ \hline
I feel \hl{sorry} for this little girl. She's legitimely scared.                                                                                       & Remorse       & \multirow{2}{*}{\xmark} & \multirow{2}{*}{\xmark} & \multirow{2}{*}{\xmark} & \multirow{2}{*}{\cmark} \\
 \\ 
\bottomrule
\end{tabular}
\caption{An overview of predictions done by different $BERT$ models on GoEmotions.}
\label{tab:examples}
\end{table*}

In this work, we perform empirical research to investigate the performance of formative FLM: BERT and its variant ($BERT_{base}$,  and  $BERT_{large}$) on two types of datasets: benchmarks and real-world. We discover substantial disagreement\footnote{one-tailed binomial t-test; p$>>$0.01)} among variants of BERT on a real-world classification dataset (see table \ref{tab:examples}). A simple solution to mitigate this problem is the ensembling of FLMs by taking into account the ``best of both the worlds'' approach. It would allow FLMs with varying capabilities to generate contextualized language representations by utilizing their relative strengths and adjusting their weaknesses. With this intuition, we present Shallow, Semi, and Deep Ensemble methods of pairing BERT. We present Deep Ensemble as a knowledge-infused reinforcement learning-based BERT ensemble strategy, which takes the support of Wikipedia and CommonSense knowledge graphs to evaluate the correctness of its prediction before yielding an outcome \cite{speer2017conceptnet}\cite{yamada2018wikipedia2vec}.
On benchmark datasets-- GoEmotions \cite{demszky2020goemotions} and Stanford Natural Language Inference (SNLI) \cite{bowman2015large}, the following Shallow, Semi, and Deep Ensembles significantly outperform BERT's variants. In addition, PRIMATE \cite{gupta2022learning} and Twitter COVID-19 datasets, which are considered to be real-world, we found similar success through ensembling. We selected these datasets with the observation that they are used to develop applications that sit on top of FLMs.

\section{METHODOLOGY}

Let $\mathcal{D}=\{x^{(i)}, y^{(i)}: i=1, \dots, m\}$ be the given dataset where $x^{(i)}$ is the feature vector and $y^{(i)}$ is the observed class for the $i$th sentence. $y^{(i)}$ can take values from 1 to $c$. Let $\mathcal{M}=\{M_{\ell} : \ell=1, \dots, n\}$ be a collection of $n$~ FLMs. We next discuss different ensemble strategies used in this paper.


\subsection{Shallow-Ensemble Strategy}

For each $x^{(i)}$, we obtain the estimated probability of it belonging to a certain category $k$ using model $M_\ell$. Denote that probability as $\text{Prob}_\ell(y^{(i)} = k | x^{(i)})$. Given some weights $\alpha_1,\dots,\alpha_n$ such that $\alpha_\ell \in [0,1]$ and $\sum_{l=1}^{n}\alpha_\ell = 1$, we combine these probabilities from different models as
$$\sum_{\ell=1}^{n}\alpha_\ell \text{Prob}_\ell(y^{(i)} = k | x^{(i)}).$$

We obtain the predicted class for the $i$th sentence as
$$ \hat{y}^{(i)} (\boldsymbol{\alpha}) = \argmax_k \sum_{\ell=1}^{n}\alpha_\ell \text{Prob}_\ell(y^{(i)} = k | x^{(i)}),$$

where $\boldsymbol{\alpha} = (\alpha_1,\dots,\alpha_n)$.

We define the loss as a function of $\boldsymbol{\alpha}$ as
$$ L (\boldsymbol{\alpha}) = \sum_{i=1}^{m} I_{[y^{(i)} \ne \hat{y}^{(i)} (\boldsymbol{\alpha})]}$$

In our Shallow-Ensemble strategy, we are interested in solving Problem \ref{problem:1}.
\begin{problem} \label{problem:1}
\begin{align*} 
    & \argmax_{\boldsymbol{\alpha}} \ L (\boldsymbol{\alpha}),\\
    \text{s.t.} \qquad & \alpha_\ell \in [0,1] \ \forall \ell, \\
    \qquad & \sum_{\ell=1}^{n} \alpha_\ell = 1. 
\end{align*}
\end{problem}

\noindent \textbf{Note.}
In this ensemble, FLMs use simple statistical method by averaging the probability output \cite{veit2016residual}. 





\subsection{Semi-Ensemble Strategy} \label{subsec:semi}


For each $x^{(i)}$, we obtain its sentence embedding using model $M_\ell$. Denote that embedding as $E_\ell(x^{(i)})$. We obtain the following ensemble embedding for $x^{(i)}$ by combining these model-level embeddings.
$$E(x^{(i)}) := [E_1(x^{(i)}), E_2(x^{(i)}),\dots, E_n(x^{(i)})].$$

We train a `smaller' model $S$ using the dataset $\tilde{\mathcal{D}}=\{E(x^{(i)}), y^{(i)}: i=1, \dots, m\}$ and obtain the estimated probability of it belonging to a certain category $k$. Denote that probability as $\text{Prob}_S(y^{(i)} = k | x^{(i)})$.

We obtain the predicted class for the $i$th sentence as
$$ \hat{y}^{(i)} (S) = \argmax_k \text{Prob}_S(y^{(i)} = k | x^{(i)}).$$

We define the loss as a function of the model $S$ as
$$ L (S) = \sum_{i=1}^{m} I_{[y^{(i)} \ne \hat{y}^{(i)} (S)]}$$

In our Semi-Ensemble strategy, we are interested in solving Problem \ref{problem:2}.
\begin{problem} \label{problem:2}
\begin{align*} 
    & \argmax_{S} \ L (S).
\end{align*}
\end{problem}

\noindent
\textbf{Note.}
In this ensemble, the total number of FLMs is fixed, but their weighted average can change. We may transform the embedding to a lower-dimensional space using standard dimensionality reduction methods like PCA, TSNE, and LSH.

\subsection{Deep-Ensemble Strategy}

For each $x^{(i)}$, we obtain knowledge graph-based embeddings using the Wikipedia \cite{yamada2020wikipedia2vec} and CommonSense \cite{speer2017conceptnet} knowledge graphs. Denote those embeddings as $E_\text{Wiki}(x^{(i)})$ and $E_\text{Comm}(x^{(i)})$, respectively. Let $E(x^{(i)})$ the ensemble embedding of $x^{(i)}$ as defined in Section \ref{subsec:semi}. Given some weight $\beta$, we calculate the following similarity metric.
\begin{align*}
w^{(i)}(\beta):= \beta \ & \text{Cosine-Sim}(E_\text{Wiki}(x^{(i)}),E(x^{(i)})) \\ 
+ (1-\beta) & \text{Cosine-Sim}(E_\text{Comm}(x^{(i)}),E(x^{(i)}))
\end{align*}

We define the following reward function.
$$R({\beta}) = \sum_{i=1}^{m} w^{(i)}(\beta) I_{[y^{(i)} = \hat{y}^{(i)}(M) (\boldsymbol{\alpha})]}$$

where $M$ can be an individual FLM, Shallow-, or Semi-Ensemble of different FLMs.

In our Deep-Ensemble strategy, we are interested in solving Problem \ref{problem:3}.
\begin{problem} \label{problem:3}
\begin{align*} 
    & \argmax_{\beta} \ R({\beta}),\\
    \text{s.t.} \qquad & \beta \in [0,1].\\
\end{align*}
\end{problem}

\begin{figure}
\centerline{\includegraphics[width=0.52\textwidth, scale=1.0]{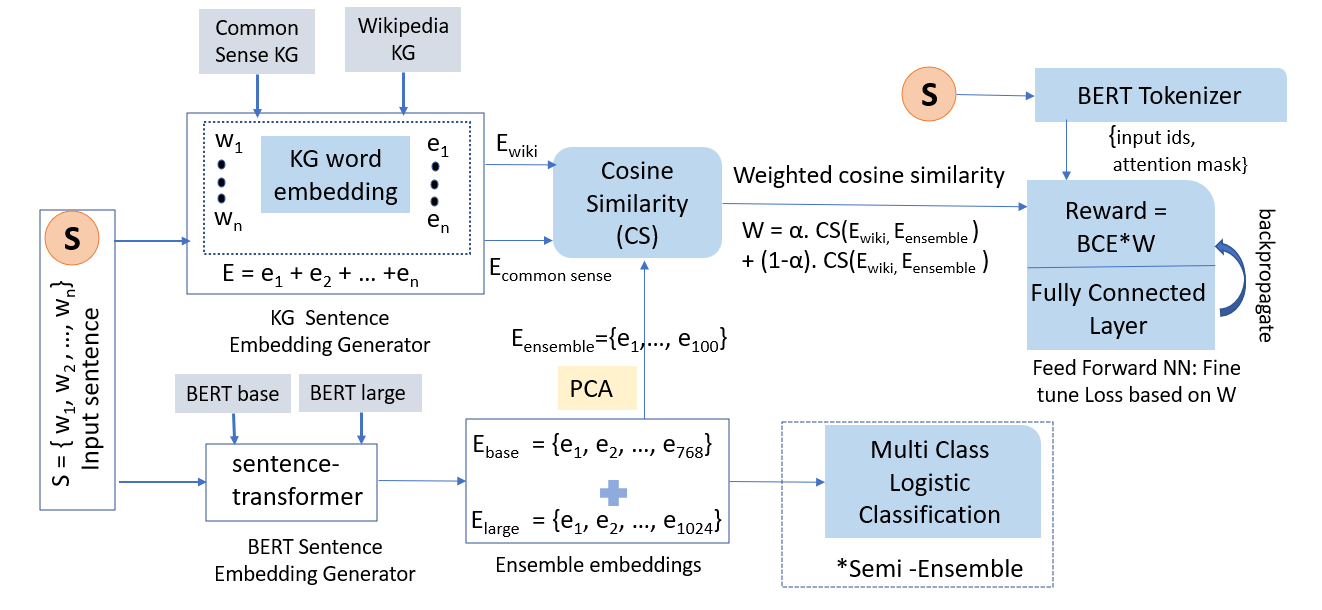}}
\caption{Illustration of the Deep-Ensemble. We train Deep-Ensemble using a policy gradient that relies on knowledge graphs(KG) as discussed in section 2.2. * denotes the Semi-Ensemble as discussed in section 2.1.}
\label{fig1}
\end{figure}

\noindent 
\textbf{Note.} The generated reward of $R(\beta)$ function is applied to the Binary Cross-Entropy Loss function of the classifier, to act as a compensator to tune the loss according to the RL Policy Gradient Method as illustrated in Figure \ref{fig1}. In terms of trainability, Deep-Ensemble has the most hyper-parameters which can be tuned to enhance the ensemble performance. Applying ground truth knowledge to the classifier through loss function helps to take best of the sentence semantic and significantly increase the performance of the ensemble compared to previous methods.

\section{Experiments}
\noindent \textbf{Datasets:} We use two standard benchmarks and two real-world datasets to evaluate the performance of FLMs and Ensembles. GoEmotions contains 58k carefully curated Reddit comments labeled for 27 emotion categories and neutral for the sentiment classification task. PRIMATE is a real-world dataset on mental health with posts coming from Reddit’s subreddit r/depression\_help. \textbf{PRIMATE} stands for \textit{PRocess knowledge Integrated Mental heAlth daTasEt}. PRIMATE aims to train conversational agents to identify parts of the user’s content that can answer a certain number of questions in clinical questionnaires like the Patient Health Questionnaire (PHQ-9). The dataset consists of $\sim2500$ posts annotated with nine ``yes'' or ``no'' labels corresponding to whether or not the post answers the nine PHQ-9 questions. \textbf{SNLI} corpus is the standard dataset for Natural Language Inferencing application, a collection of 570K labeled human-written English sentence pairs. \textbf{Twitter} dataset is tasked to train machine learning and deep learning models for sentiment detection on 1.6 Million tweets collected during COVID-19 Pandemic. Each row in the dataset is marked as 1 for positive and 0 for negative. 

\begin{table*}[!ht]
\footnotesize

    \centering
    \begin{tabular}{p{1.2cm}p{1.5cm}p{1.5cm}p{1.5cm}p{1.5cm}|p{1.cm}p{1cm}p{1.cm}|p{1cm}|p{0.5cm}}
        \toprule[1.5pt]
         Dataset & $BERT_{tiny}$ & $BERT_{mini}$ & $BERT_{base}$ & $BERT_{large}$ & ShE & SE & DE & XLNet & T5  \\ \midrule
         GoEmotions$\dagger$ & 3.9 & 4.2 & 5.1 & 3.4 & 5.2 & 35.8 & \textit{49.7} & 48.2 & \textbf{50.6}   \\
         PRIMATE$\ddagger$ & 57.2 & 60.1 & 60.4 & 43.5 & 64.5 & \textbf{75.8} & \underline{70.5} & 69.2 & \textbf{72.7} \\
         SNLI$\dagger$ & 88.2 & 85.3 & 92.0 & 91.6 & 92.7 & 85.3 & \textbf{99.1} & 96.8 & 97.2 \\
         Twitter$\ddagger$ & 29.8 & 34.4 & 41.6 & 51.0 & 63.6 & 42.5 & \underline{65.0} & \textbf{67.4} & 66.5 \\ \bottomrule[1.5pt]
    \end{tabular}
    \caption{Performance of single FLMs (BERT, XLNET, T5), ensembles of BERT without knowledge (ShE: Shallow-Ensemble; SE: Semi-Ensemble) and Deep-Ensemble~(DE) of BERT with knowledge on the four datasets. In majority of cases, ensembles has \textbf{best}, \textit{second best performance}, or \underline{close performance} to larger FLMs. $\dagger$ resembles statistically significant performance at $p<0.001$ and  $\ddagger$ resembles significance at $p<0.05$ using one-tailed binomial t-test.}
    \label{tab:my_label}
\end{table*}

\noindent\textbf{Results and Discussion:} We report the performance of ShE, SE, and DE with $BERT_{tiny}$, $BERT_{mini}$, $BERT_{base}$, and $BERT_{large}$ in table \ref{tab:my_label} \cite{devlin2018bert}\cite{jiao2019tinybert}.

\noindent \textbf{\textit{What is the impact of shallow ensembles (ShE)?} }We note that ShE showed an improvement of 19.9\%, 14.2\%, 3.42\%, and 38.34\% percentage improvement over the single FLMs on GoEmotions, PRIMATE, SNLI, and Twitter respectively. To better understand the performance gains, consider the sentence (\textbf{S})~\textit{please calm down}, $BERT_{large}$ laid its attention on \textit{calm}, $BERT_{base}$ attention was on \textit{calm down}. Then ShE of $BERT_{large}$ and $BERT_{base}$ gave relatively high attention weight to the word \textit{please}, along with the phrase \textit{calm down}, showing the true label: \textit{annoyance}, with high probability. Even by changing \textbf{S} from \textit{please calm down} to \text{please calm down when feeling low}, the attention weights of ShE were more substantial on \textit{please, calm down} and \textit{feeling}.

\noindent \textbf{\textit{What is the impact of Semi-Ensembles (SE)?}} The following ensembles-- SE, ShE and DE underperforms on SNLI and Twitter datasets and outperforms on GoEmotions and PRIMATE datasets by 85.47\% and 14.9\% respectively.  
To compare ShE and SE, we take a sentence from PRIMATE. Let \textbf{S2} be \textit{life is crumbling to pieces}, where SE correctly predicts the class label as \textit{no}. The attention weights were more on the words \textit{crumbling and pieces} as compared to ShE where the attention was more focused on \textit{life}. 

\noindent \textbf{\textit{What is the impact of deep ensembles (DE)?}}

For the intepretation of the attention in ShE, SE, DE, BERT, and its variants we used the tool developed by \cite{chefer2021transformer}. With respect to ShE and SE, DE outperforms both of them on GoEmotions, SNLI and Twitter by 27.9\%, 13.9\%, and 34.61\% respectively.  It underperforms on PRIMATE. To compare SE and DE, we take a sentence from GoEmotions. Let \textbf{S3} be \textit{Now you ruined the surprise} DE laid greater attention on the words \textit{ruined and surprised}, showing the true label \textit{neutral} with high probability. For \textbf{S3}, SE laid greater attention to \textit{surprised} and incorrectly predicts the true label. 


\section{Conclusion}


The current research presented ensembles of FLMs to perform an empirical reality check. Specially, we highlighted the spurious and brittle behavior of single-running FLMs, which can be resolved through an ensemble. In this direction, we presented three ensemble categories and conducted experimentation of benchmarking and real-world datasets. First, we evaluated how ensembling multiple language models (e.g., the BERT model) with different ensemble techniques could generate more accurate, efficient, and robust results for natural language sentence classification tasks. We evaluated the models with other standard benchmarks and real-world datasets, and the experimental results demonstrate that all ensembling methods, even the naive ones, outperform the traditional baseline language models and demonstrate considerable superiority in prediction accuracy. We also observe that ensembling FLMs with a classifier whose loss is tuned by the reinforcement learning method and infused by typical human knowledge graphs leads to significant improvements in performance.  

\paragraph{Reproduciblity and Impact:} While ensembles are well-known in machine learning, they aren't considered the core building blocks in FLMs or develop effective models. On the other hand, ensembles discussed in the current research are simple to build and easy to deploy on resource-constrained edge or mobile devices \cite{gunaratna2021using}. For instance, the PRIMATE dataset's purpose was to enhance the capability of FLMs in mobile chatbots to ask effective follow-up questions while conversing with users with mental health conditions. Further, ensemble models can handle unfamiliar settings and improve adversarial robustness. We consider this as future work, along with exercising the effectiveness and limitations of the proposed deep ensemble in conversational systems.  
The implementation for \{Shallow Ensemble, Semi Ensemble, and Deep Ensemble \} is provided in the following \href{}{\textcolor{red}{\href{https://github.com/qqqxyzqqq/acl2023}{GitHub} }}. The datasets used in this code are publicly available however, we will provide a processed version (e.g. after removing slangs) on the same github link.

\appendix
\section{Limitations and Ethical Statement}
To highlight the effectiveness of ensembles, we intentionally choose classification tasks, where attention of the FLMs can be explained. Further, we did not include certain other ensemble models like PABEE and LeeBERT for our work \cite{xu2022survey}. These models have been used as early exit models to prevent FLMs from underfitting or overfitting. 

We focused on the value of user privacy and the delicate nature of social media while analyzing the PRIMATE and Twitter COVID-19 datasets. As a result, we only use datasets that already exist, and all user data is stored separately on secure servers and only indirectly connected to raw text and network data via anonymous IDs \cite{bruckman2002studying}\cite{chancellor2016quantifying}.




\bibliographystyle{acl_natbib}
\bibliography{ensemble_BERT_update_1}
\end{document}